\documentclass{article}
\usepackage{spconf,amsmath,graphicx}
\usepackage{array}
\usepackage{color}
\usepackage{version}
\usepackage{caption}
\usepackage{url}
\usepackage{tikz}

\newcommand{\SSS}{\mathcal{S}}

\definecolor{gg}{RGB}{0,100,0}

\title{Robust Shape Regularity Criteria for Superpixel Evaluation}
\name{R{\'e}mi Giraud$^{1,2}$ \qquad Vinh-Thong Ta$^{1,3}$ \qquad Nicolas Papadakis$^{2}$
\thanks{This work has been carried out with financial support of the French 
State, managed by the French National Research Agency (ANR) in the  
frame of the GOTMI project (ANR-16-CE33-0010-01) and
the Investments for the future Program IdEx Bordeaux 
(ANR-10-IDEX-03-02) with the Cluster of excellence CPU.
}
}
\address{$^{1}$Univ. Bordeaux, LaBRI, CNRS, UMR 5800, F-33400 Talence, France.\\
    $^{2}$Univ. Bordeaux, IMB, CNRS, UMR 5251, F-33400 Talence, France.\\
     $^{3}$Bordeaux INP, LaBRI, UMR 5800, F-33405 Talence, France.
}    
    
\begin{document}
%
\maketitle
\begin{abstract}

Regular decompositions are necessary
for most superpixel-based object recognition or tracking applications.
So far in the literature, the regularity or compactness of a superpixel shape
is mainly measured by its circularity.
In this work, we first demonstrate that such measure
is not adapted for superpixel evaluation, since it does not directly express regularity
but circular appearance.
Then, we propose a new metric that
considers several shape regularity aspects:
convexity, balanced repartition, and contour smoothness.
Finally, we demonstrate that our measure
is robust to scale and noise and 
 enables to more relevantly compare 
superpixel methods.

\end{abstract} \medskip

\noindent

\section{Introduction}
\label{sec:intro}

The decomposition of an image into homogeneous areas, called \textit{superpixels},
has become a very popular pre-processing step in many image processing and computer vision 
frameworks.
For most superpixel-based applications such as 
object recognition \cite{tighe2010,gould2014}, tracking \cite{wang2011,chang2013} or labeling \cite{giraud2017},
the use of regular decompositions
\cite{achanta2012,machairas2015,li2015,yao2015,zhang2016,giraud2016}
where superpixels roughly have the same size and regular shapes is necessary.
With such regularity, accurate superpixel features can be computed
and relevant information can be extracted from their boundaries, while
with irregular decompositions \cite{felzenszwalb2004,liu2011,vandenbergh2012,buyssens2014,rubio2016},
superpixels can have different sizes, noisy boundaries and stretched shapes.

Most of recent methods 
allow the user to tune a compactness parameter \cite{achanta2012,buyssens2014,machairas2015,li2015,yao2015,giraud2016}
to produce superpixels of variable regularity, that may affect the 
performances for a given application.
The search for optimal results and comparison between methods should 
thus be performed for several regularity settings \cite{schick2012}.
Moreover, most superpixel decomposition
methods compute a trade-off between
adherence to contours and shape compactness.
The regularity is hence an argument to compare decompositions
with similar contour adherence performances.
Hence, clear regularity definition and measure are necessary to 
evaluate superpixel methods.

\begin{figure}[t!]
\begin{center}
{\scriptsize
\begin{tabular}{@{\hspace{0mm}}c@{\hspace{5mm}}c@{\hspace{0mm}}}
\hspace{-0.8cm} {\scriptsize C = $0.843$ $|$ \textbf{SRC = $\mathbf{0.961}$}}&
\hspace{-0.8cm} {\scriptsize C = $0.890$ $|$ \textbf{SRC = $\mathbf{0.932}$}}\\
 \begin{tikzpicture}
\node[anchor=north west,inner sep=0pt,outer sep=0pt] at (0,0) {\includegraphics[width=0.21\textwidth,height=0.10\textwidth]{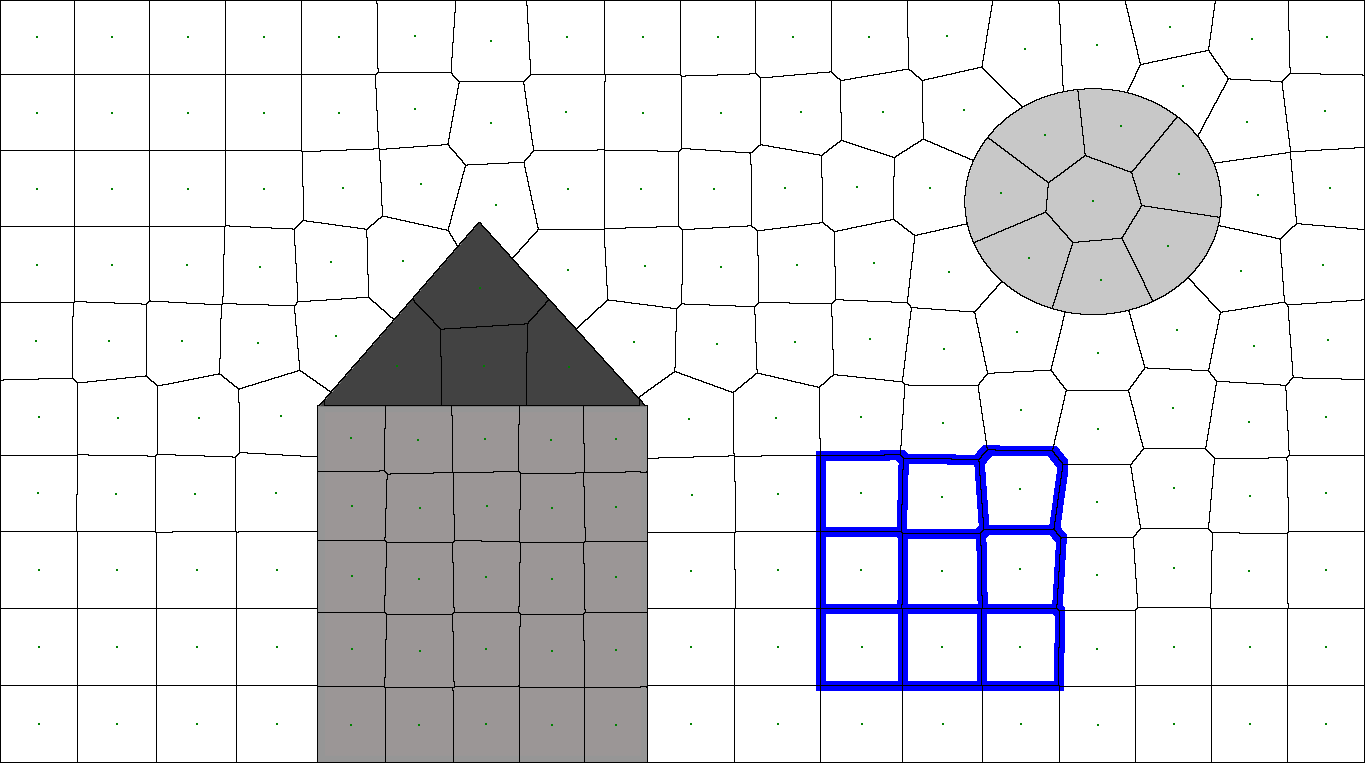}};
\draw [->] [blue] (2.6,-1.65) -> (2.5,-1.85);
\end{tikzpicture}&
 \begin{tikzpicture}
\node[anchor=north west,inner sep=0pt,outer sep=0pt] at (0,0) {\includegraphics[width=0.21\textwidth,height=0.10\textwidth]{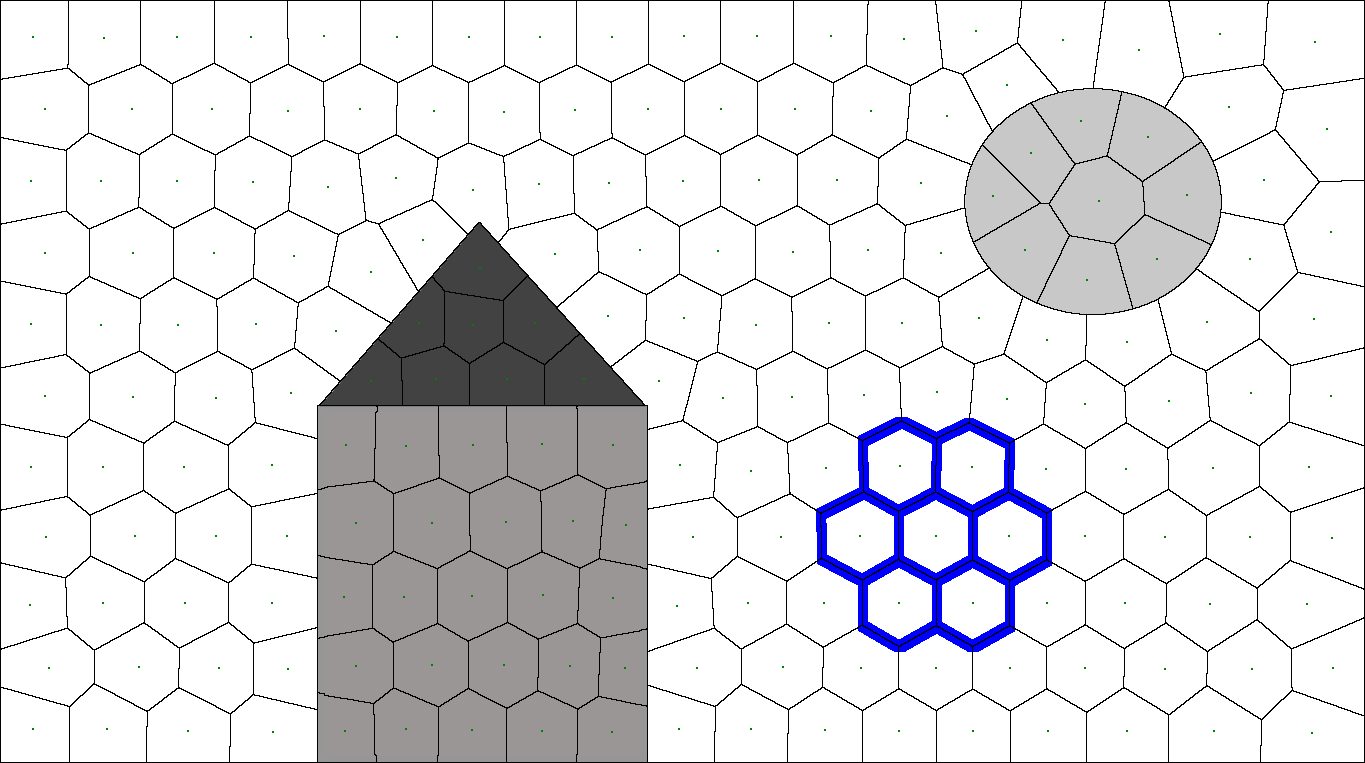}};
\draw [->] [blue] (2.55,-1.55) -> (2.5,-1.85); 
\end{tikzpicture} \\  
\hspace{0.8cm}  {\scriptsize {\color{blue} C = $0.850$ $|$ \textbf{SRC = $\mathbf{0.974}$}} }&
\hspace{0.8cm} {\scriptsize {\color{blue}  C = $0.951$ $|$ \textbf{SRC = $\mathbf{0.968}$}} }
 \end{tabular}
 }
\end{center}
\caption{Two decompositions
with square (left) and hexagonal (right) initialization using  \cite{achanta2012}.
The
circularity C is much higher for hexagons,
and for the whole right image, despite the stretched hexagonal shapes at the
borders,
while the proposed SRC equally evaluates square and hexagon and
provides more relevant regularity measure.
} \vspace{-0.4cm}
\label{fig:init}
\end{figure}

In the superpixel literature, the regularity notion is usually called \textit{compactness}, 
and is mainly evaluated by the circularity metric \cite{schick2012}.
Although the circularity metric has since been considered
in many works 
\cite{reso2013,buyssens2014,zhang2016,giraud2016}, and 
large benchmarks \cite{schick2014,stutz2016},
the literature usually refers to the regularity
as the ability to produce convex shapes with non noisy boundaries.
Moreover, for tracking application, the aim is to find
one-to-one superpixel associations across decompositions,
so compactness should be high for convex shapes with balanced pixel repartition.
In \cite{machairas2015}, the circularity is also discussed since it
does not consider the square as a highly regular shape.
Figure \ref{fig:init} compares two decompositions computed with \cite{achanta2012}
using initial square and hexagonal grids.
Hexagons provide much higher circularity, although 
both shapes should be considered as regular.
Since most methods start from a square grid
and iteratively refine the superpixel borders,
it would make sense to have high regularity for a square decomposition,
and to have a measure that is in line with the 
compactness parameter evolution,
that produces squares when set to maximum value.

\noindent\textbf{Contributions.}
In this work, we first demonstrate that the circularity is not adapted to the
superpixel context.
We propose a new shape regularity criteria (SRC), that 
better expresses the regularity notion by
considering the following aspects:
shape convexity, balanced pixel repartition and contour smoothness.
Finally, the relevance and the robustness of SRC are demonstrated
with state-of-the-art superpixel methods applied to
images of the Berkeley segmentation dataset \cite{martin2001}.

\section{Shape Regularity Criteria}
\label{sec:shape}

\subsection{Regularity Definition}

The circularity C,
introduced in \cite{schick2012} for superpixel evaluation,
expresses the compactness of a shape $S$
as follows: \vspace{-0.1cm}
\begin{equation}
\text{C}(S) = 4\pi |S|/|P(S)|^2 ,
\end{equation}
where $P(S)$ is the shape perimeter and
$|.|$ denotes the cardinality.
This metric considers the compactness
as the resolution of the 
isoperimetrical problem that aims to find the largest shape
for a given boundary length.
As stated in the introduction, 
the circularity does not express the shape regularity but only favor circular shapes.
We propose a new shape regularity criteria (SRC)  
 composed of three metrics,
that each evaluates 
an aspect of the superpixel regularity. \smallskip

\noindent \textbf{Solidity.}

\begin{figure}[t!]
\begin{center}
{\small
\begin{tabular}{@{\hspace{0mm}}c@{\hspace{5mm}}c@{\hspace{5mm}}c@{\hspace{0mm}}}
 \includegraphics[width=0.11\textwidth]{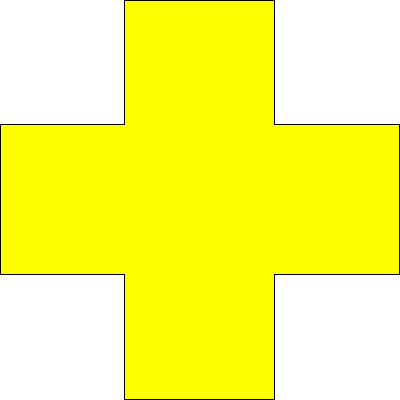} &  
 \includegraphics[width=0.11\textwidth]{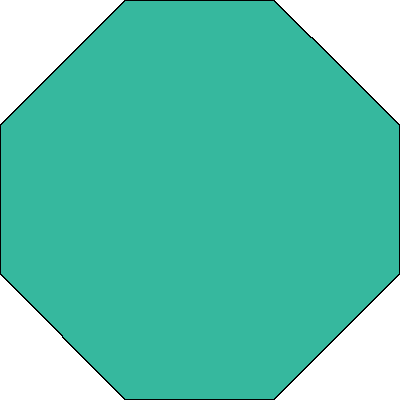}&
 \includegraphics[width=0.11\textwidth]{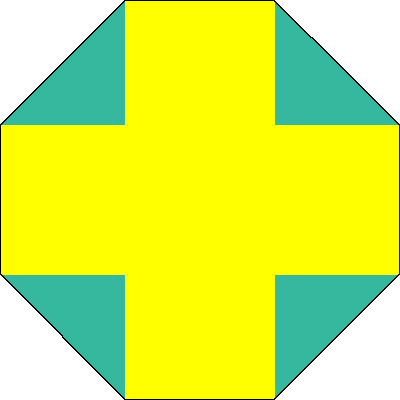}\\
  (a)&(b)&(c)
 \end{tabular}
 }
\end{center}
\caption{Convex hull example on a synthetic shape.
The overlap between the shape (a) and its convex hull (b)
is shown in (c). The shape is contained into the hull and
the overlap is such that $\text{SO}=78\%$.} \vspace{-0.4cm}
\label{fig:hull}
\end{figure}

To evaluate the global convexity of a shape,
we propose to consider its \textit{solidity} (SO), \emph{i.e.}, 
the overlap with its convex hull $CH$.
Such convex hull,
containing the whole shape $S$,
is illustrated in Figure \ref{fig:hull} and
can be computed using Delaunay triangulation.
Perfectly convex shapes such as squares or circles will get the highest
\textit{solidity}: \vspace{-0.1cm}
\begin{equation}
\text{SO}(S) =  |S|/|CH|  \leq 1 .
\end{equation}

\noindent\textbf{Balanced repartition.} 
The overlap with the convex hull is not sufficient to 
express the global regularity. 
Convex shapes such as ellipses or lines have the highest SO,
but should be
considered as perfectly regular only with a
balanced pixel repartition.
To measure it, we
define a variance term $\text{V}_{\text{xy}}$: \vspace{-0.1cm}
\begin{equation}
\text{$\text{V}_{\text{xy}}$}(S) =  \sqrt{\min(\sigma_x,\sigma_y)/\max(\sigma_x,\sigma_y)}  \leq 1 ,
  \end{equation}
\noindent with $\sigma_x$ and $\sigma_y$, the standard deviations of
the pixel positions
$x$ and $y$ within $S$. 
$\text{$\text{V}_{\text{xy}}$} = 1$ if, and only if, $\sigma_x=\sigma_y$.
In this case, 
the spatial repartition of the pixels around the barycenter is considered as well balanced.
\smallskip

\noindent\textbf{Contour smoothness.}
Finally, the regularity of the superpixel borders
must be considered.
The \textit{convexity} measure (CO) compares the
number of boundary pixels of the shape and the one of its convex hull.
Although
this measure is generally in line with SO, it
is mostly dependent on the 
border smoothness and
penalizes noisy superpixels:
\begin{equation}
 \text{CO}(S) =  |P(CH)|/|P(S)|   \leq 1 .
\end{equation}

The proposed shape regularity criteria (SRC)
is a combination of all regularity aspects
and is defined
as follows: \vspace{-0.1cm}
\begin{equation}
\text{SRC}(\SSS) = \sum_k{\frac{|S_k|}{|I|}\text{SO}(S_k)}\text{$\text{V}_{\text{xy}}$}(S_k)\text{CO}(S_k) , 
\end{equation}
where  $\mathcal{S}=\{S_k\}_{k\in \{1,\dots ,|\mathcal{S}|\}}$ is
composed of  $|\mathcal{S}|$ superpixels $S_k$, whose sizes are considered
to reflect the overall regularity.
The combination of SO and CO is related to the ratio between
the Cheeger constant of the shape $S$ and its convex hull $CH$.
As stated in \cite{caselles2009}, the Cheeger measure is
a pertinent tool to evaluate
both convexity and boundary smoothness.

\subsection{Circularity vs SRC}
Figure \ref{fig:shape} compares the different metrics
on synthetic shapes split into three groups, and generated
with smooth (top) and noisy borders (bottom).
The circularity presents several drawbacks.
First, it
is much lower for the \textit{Square} than for the \textit{Circle} and \textit{Hexagon}, 
and even less than for the \textit{Ellipse}.
As shown in Figure \ref{fig:init}, this is an issue when comparing superpixel  methods
starting from square and hexagonal grids, such as \cite{machairas2015}.
It also drops for the \textit{Cross} and \textit{Bean} although they are visually regular.
In the bottom part of Figure \ref{fig:shape},
the circularity appears to be very dependent on the boundary smoothness,
since noisy shapes have similar circularity and cannot be differentiated.
Finally, standard shapes with smooth borders 
can have much higher circularity than regular ones.
For instance, the noisy \textit{Square} has a lower circularity than the \textit{Bean}.

As can be seen in Figure \ref{fig:shape}, 
SO, $\text{V}_{\text{xy}}$ and CO independently taken  are not sufficient 
to express the compactness of a shape. 
The proposed SRC
combines all defined
regularity properties.
For instance, 
SO is representative for all shapes, except for the \textit{Ellipse} and \textit{W},
since they both have large overlap with their convex hull.
$\text{V}_{\text{xy}}$ penalizes the \textit{Ellipse} 
since it does not have a balanced pixel repartition, and
CO considers the large amount of contour pixels in the \textit{W} shape.

\begin{figure}[t!]
\begin{center}
\begin{tabular}{@{\hspace{0mm}}c@{\hspace{0mm}}}
  \includegraphics[width=0.23\textwidth,height=0.18\textwidth]{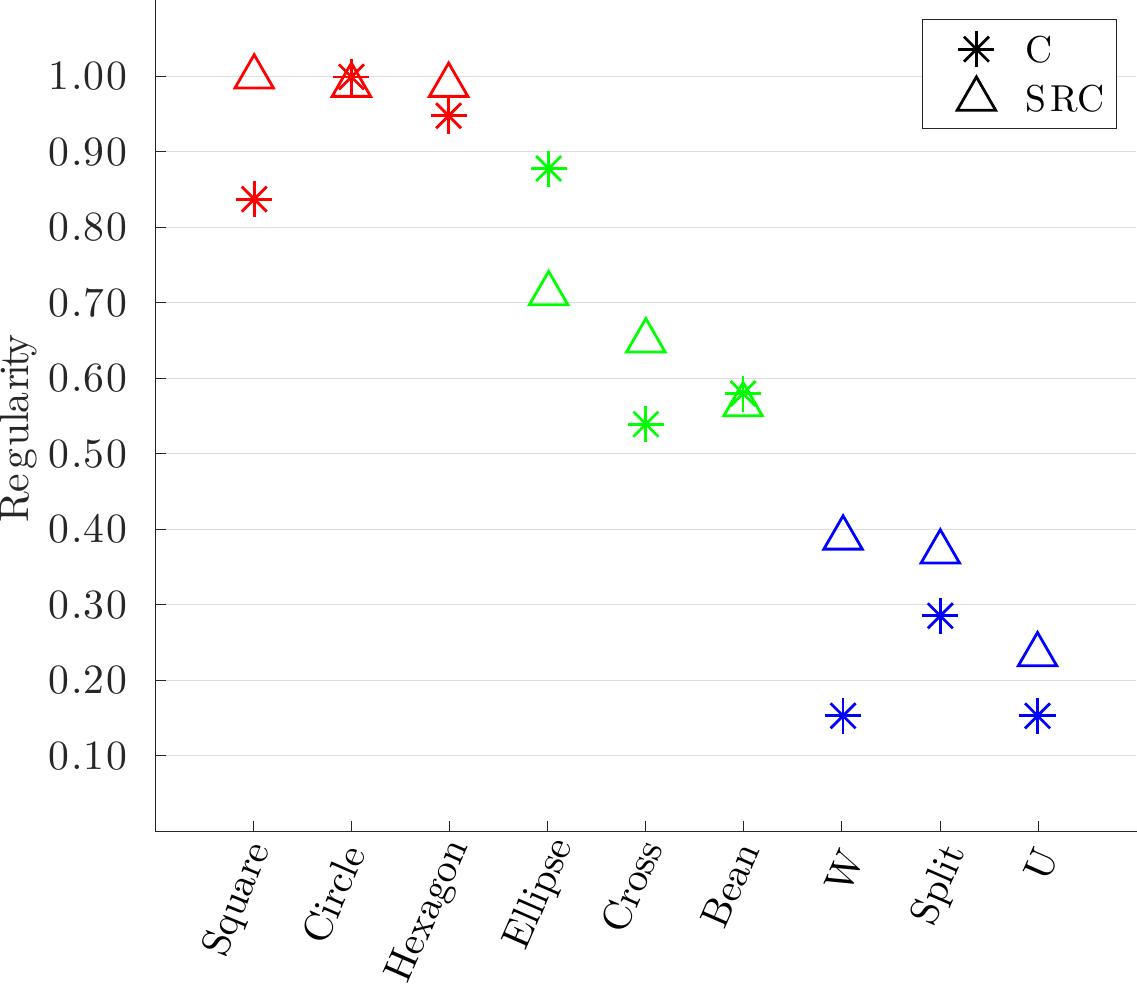}
    \includegraphics[width=0.23\textwidth,height=0.18\textwidth]{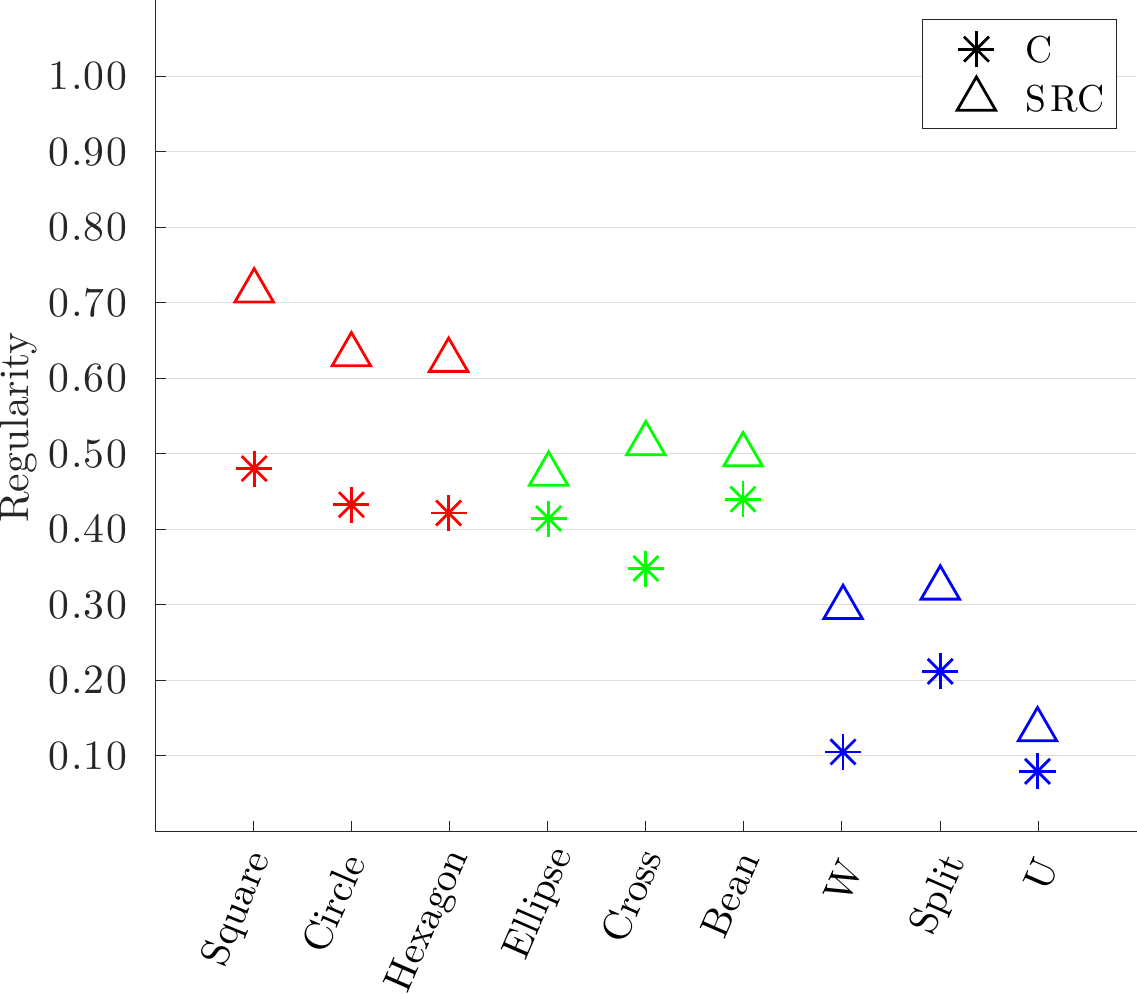} \\
 \end{tabular}
\end{center}
\caption{C and SRC on smooth (left) and noisy shapes (right).
SRC more clearly separates in both cases the three shape groups
(regular in red, standard in green and irregular in blue).
} \vspace{-0cm}
\label{fig:c_src}
\end{figure}

The three regular shapes get the highest SRC ($\approx1$), and 
the
standard shapes have similar measures.
Since our metric is
less sensitive to slight contour smoothness variations,
SRC also clearly separates the three shape groups
in the noisy case, contrary to C.
Moreover, regular but noisy shapes,
have comparable SRC to smooth standard ones,
whereas noisy standard shapes
still have higher 
SRC than irregular shapes with smooth contours,
which
can be considered as 
a relevant evaluation of regularity.
Figure \ref{fig:c_src} also represents the regularity measures, where
SRC appears to more clearly separates the three shape groups in both smooth and noisy cases.

\newcommand{\ec}{4mm}
\newcommand{\shapew}{0.06\textwidth} 
\begin{figure*}
\begin{center}
{\footnotesize
 \begin{tabular}{@{\hspace{0mm}}p{1.5cm}@{\hspace{\ec}}c@{\hspace{\ec}}c@{\hspace{\ec}}c@{\hspace{\ec}}|@{\hspace{\ec}}c@{\hspace{\ec}}c@{\hspace{\ec}}c@{\hspace{\ec}}|@{\hspace{\ec}}c@{\hspace{\ec}}c@{\hspace{\ec}}c@{\hspace{0mm}}} 
 &\multicolumn{3}{c}{\hspace{-0.4cm}{\small Regular shapes}}&\multicolumn{3}{c}{\hspace{-0.6cm}{\small Standard shapes}}&\multicolumn{3}{c}{{\small Irregular shapes}}  \vspace{-0.3cm} \\
 &&&&&&&&&\\ 
 	&\textit{Square}  &\textit{Circle} 	&\textit{Hexagon} 	&\textit{Ellipse} 	&\textit{Cross}	 &\textit{Bean}	  &\textit{W}	&\textit{Split}	&\textit{U} \\  
&
 \includegraphics[width=\shapew]{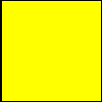}&
  \includegraphics[width=\shapew]{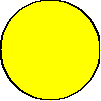}&
  \includegraphics[width=\shapew,height=\shapew]{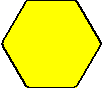}&
  \includegraphics[width=\shapew]{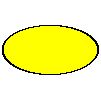}&
  \includegraphics[width=\shapew]{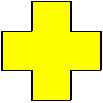}&
  \includegraphics[width=\shapew]{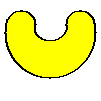}&
    \includegraphics[width=\shapew]{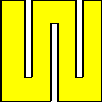}&
  \includegraphics[width=\shapew]{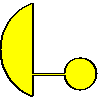}&
    \includegraphics[width=\shapew]{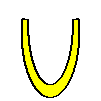}\\
\textbf{C}		&$\mathbf{0.830}$	& $\mathbf{1.000}$	& $\mathbf{0.940}$& $\mathbf{0.870}$	& $\mathbf{0.530}$	& $\mathbf{0.580}$ &
$\mathbf{0.150}$& $\mathbf{0.280}$ & $\mathbf{0.150}$ \\
\textbf{SRC}	&$\mathbf{1.000}$	& $\mathbf{0.989}$	& $\mathbf{0.987}$& $\mathbf{0.712}$	& $\mathbf{0.650}$	& $\mathbf{0.564}$ &
$\mathbf{0.387}$& $\mathbf{0.369}$ & $\mathbf{0.233}$ \\  \hline %
{\footnotesize SO} 	&${1.000}$	& ${0.989}$	& ${0.989}$& ${0.988}$	& ${0.781}$	& ${0.800}$ & ${0.841}$& ${0.530}$ & ${0.357}$ \\ 
{\footnotesize$\text{V}_{\text{xy}}$} 	&${1.000}$	& ${1.000}$	& ${0.997}$& ${0.718}$	& ${1.000}$	& ${0.811}$ & ${0.990}$& ${0.888}$ & ${0.942}$ \\ 
{\footnotesize CO} 	&${1.000}$	& ${1.000}$	& ${1.000}$& ${0.997}$	& ${0.833}$	& ${0.868}$ & ${0.465}$& ${0.783}$ & ${0.694}$ \\  \vspace{0.01cm}      
\end{tabular}
} 

\vspace{-0.2cm}

{\footnotesize
 \begin{tabular}{@{\hspace{0mm}}p{1.5cm}@{\hspace{\ec}}c@{\hspace{\ec}}c@{\hspace{\ec}}c@{\hspace{\ec}}|@{\hspace{\ec}}c@{\hspace{\ec}}c@{\hspace{\ec}}c@{\hspace{\ec}}|@{\hspace{\ec}}c@{\hspace{\ec}}c@{\hspace{\ec}}c@{\hspace{0mm}}}
 &\includegraphics[width=\shapew]{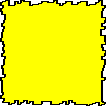}&
  \includegraphics[width=\shapew]{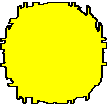}&
  \includegraphics[width=\shapew,height=\shapew]{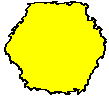}&
  \includegraphics[width=\shapew]{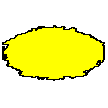}&
  \includegraphics[width=\shapew]{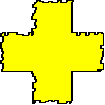}&
  \includegraphics[width=\shapew]{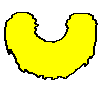}&
    \includegraphics[width=\shapew]{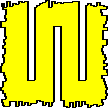}&
  \includegraphics[width=\shapew]{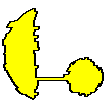}&
    \includegraphics[width=\shapew]{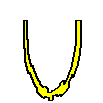}\\
\textbf{C}		&$\mathbf{0.480}$	& $\mathbf{0.430}$	& $\mathbf{0.420}$& $\mathbf{0.440}$	& $\mathbf{0.360}$	& $\mathbf{0.450}$ &
$\mathbf{0.100}$& $\mathbf{0.210}$ & $\mathbf{0.070}$ \\
\textbf{SRC}	&$\mathbf{0.716}$	& $\mathbf{0.633}$	& $\mathbf{0.625}$& $\mathbf{0.498}$	& $\mathbf{0.522}$	& $\mathbf{0.496}$ &
$\mathbf{0.296}$& $\mathbf{0.307}$ & $\mathbf{0.137}$ \\  \hline %
{\footnotesize SO} 	&${0.925}$	& ${0.923}$	& ${0.917}$& ${0.931}$	& ${0.743}$	& ${0.797}$ & ${0.763}$& ${0.542}$ & ${0.234}$ \\ 
{\footnotesize$\text{V}_{\text{xy}}$} 	&${0.999}$	& ${1.000}$	& ${0.997}$& ${0.717}$	& ${0.996}$	& ${0.802}$ & ${0.988}$& ${0.855}$ & ${0.939}$ \\ 
{\footnotesize CO} 	&${0.774}$	& ${0.685}$	& ${0.683}$& ${0.997}$	& ${0.705}$	& ${0.777}$ & ${0.392}$& ${0.662}$ & ${0.622}$ \\
\end{tabular}
}

\end{center}
\caption{Comparison of regularity metrics on 
 synthetic shapes with smooth (top) and noisy borders (bottom).
The circularity C only favors circular appearance and
does not enable to differentiate regular and standard noisy shapes.
The SRC metric
tackles these issues and more clearly separates the three shape groups 
in both smooth and noisy cases.
See text for more details.
} \vspace{-0.3cm}
\label{fig:shape} 
\end{figure*}

\begin{figure}[t!]
\begin{center}
\begin{tabular}{@{\hspace{0mm}}c@{\hspace{0mm}}}
  \includegraphics[width=0.45\textwidth,height=0.22\textwidth]{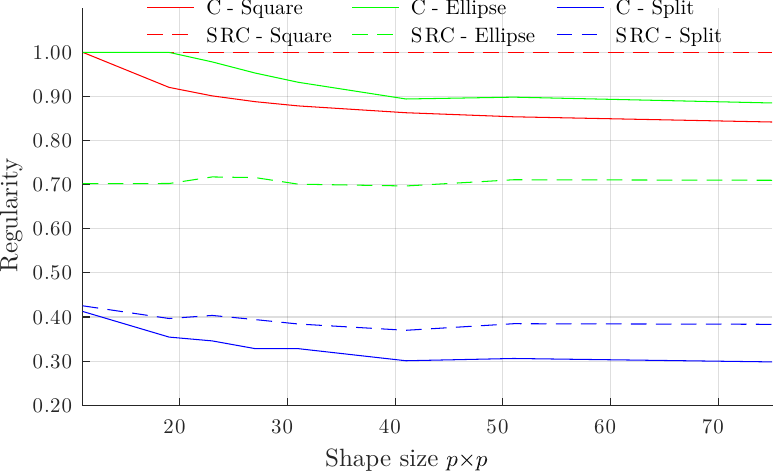} \\
 \end{tabular}
\end{center}
\caption{Comparison of circularity (C) and proposed shape regularity criteria (SRC)  
on shapes of various pixel sizes $p$.
} \vspace{-0cm}
\label{fig:size}
\end{figure}

Finally,
circularity appears to very dependent on the shape size in Figure \ref{fig:size}.
As observed in \cite{roussillon2010}, due to discrete computation, 
it can be superior to 1 (we threshold its value in Figure \ref{fig:size}), and
it drops with larger shapes.
Hence, comparisons of methods on this metric would be 
relevant only with decompositions having the same superpixel number, 
\emph{i.e.}, superpixels with approximately the same size.
Contrary to circularity, SRC provides much more consistent measure 
according to the superpixel size, \emph{e.g.}, the \textit{Square} always has a SRC equal to 1.

\section{Improved Evaluation of Superpixel Methods}
\label{sec:comparison}

\subsection{Validation Framework}

To compare results of state-of-the-art methods,
we consider the standard Berkeley segmentation dataset (BSD) \cite{martin2001},
containing 200 test images of $321{\times}481$ pixels.
At least 5 human segmentations are provided
to compute evaluation metrics of contour adherence and respect of image objects.

    \newcommand{\htab}{0.11\textwidth}
        \newcommand{\wtab}{0.192\textwidth}
  \begin{figure*}[t!]
\begin{center} 
{\footnotesize
 \begin{tabular}{@{\hspace{0mm}}c@{\hspace{1mm}}c@{\hspace{1mm}}c@{\hspace{1mm}}c@{\hspace{1mm}}c@{\hspace{0mm}}}
{ \footnotesize C = $0.473$ $|$ SRC = $0.550$}&
{ \footnotesize C = $0.333$ $|$ SRC = $0.395$}&
{ \footnotesize C = $0.547$ $|$ SRC = $0.622$}&
{ \footnotesize C = $0.238$ $|$ SRC = $0.330$}&
{ \footnotesize C = $0.587$ $|$ SRC = $0.644$}\\ 
 \includegraphics[width=\wtab,height=\htab]{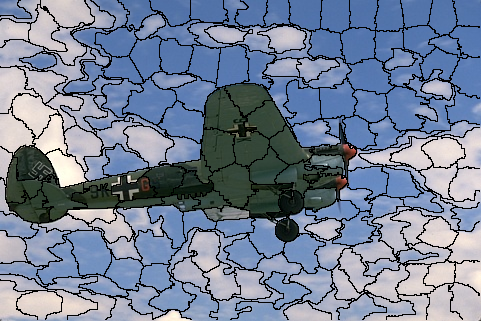}&
 \includegraphics[width=\wtab,height=\htab]{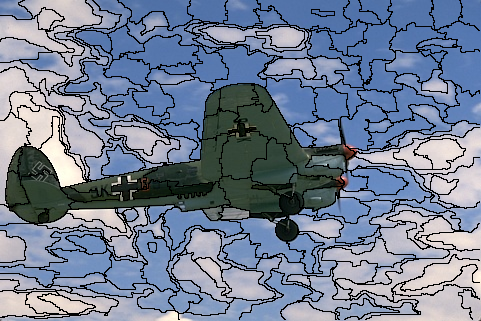}&
  \includegraphics[width=\wtab,height=\htab]{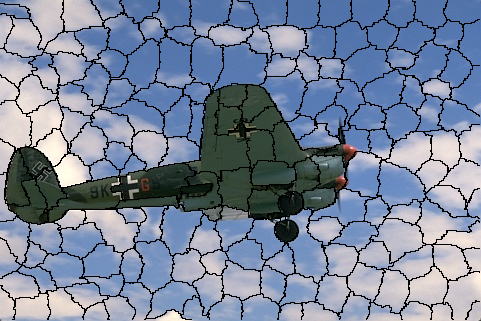}&
   \includegraphics[width=\wtab,height=\htab]{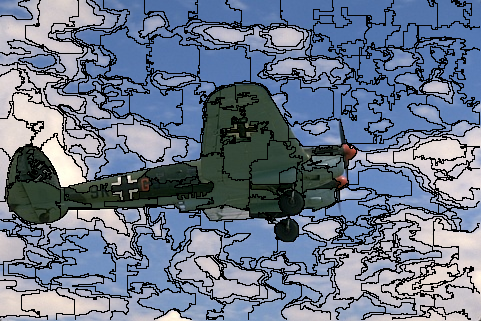}&
   \includegraphics[width=\wtab,height=\htab]{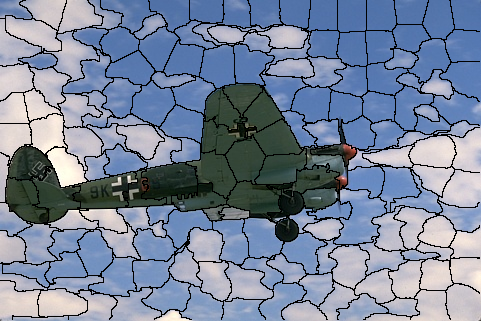}\\ 
    \includegraphics[width=\wtab,height=\htab]{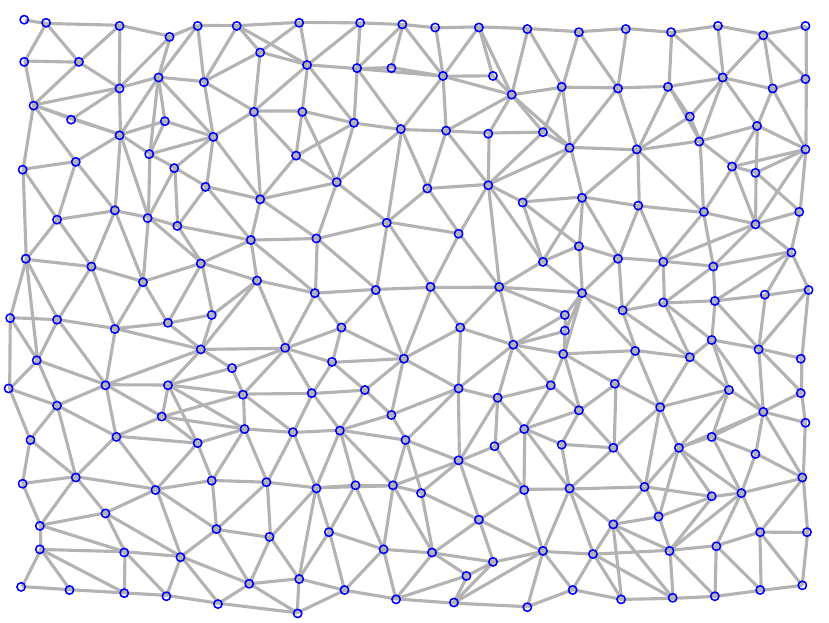}&
 \includegraphics[width=\wtab,height=\htab]{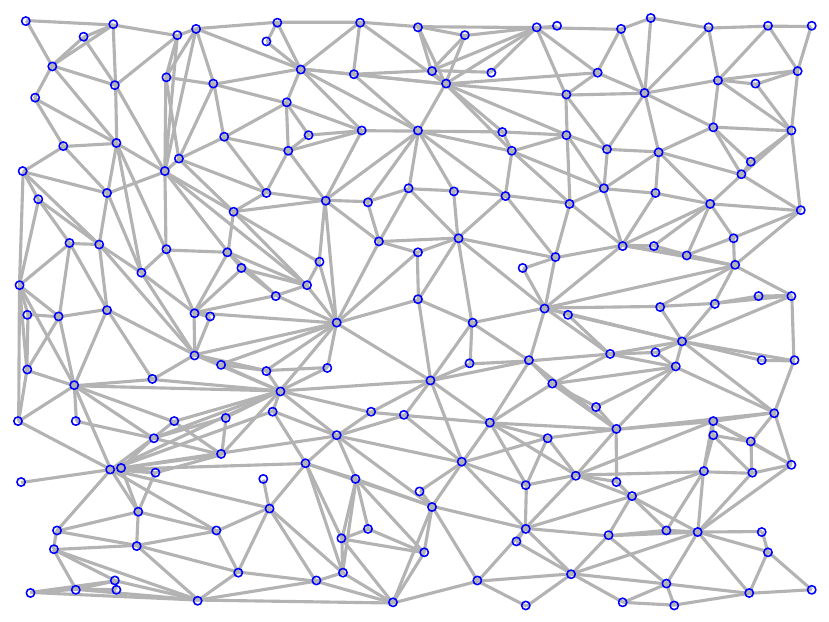}&
  \includegraphics[width=\wtab,height=\htab]{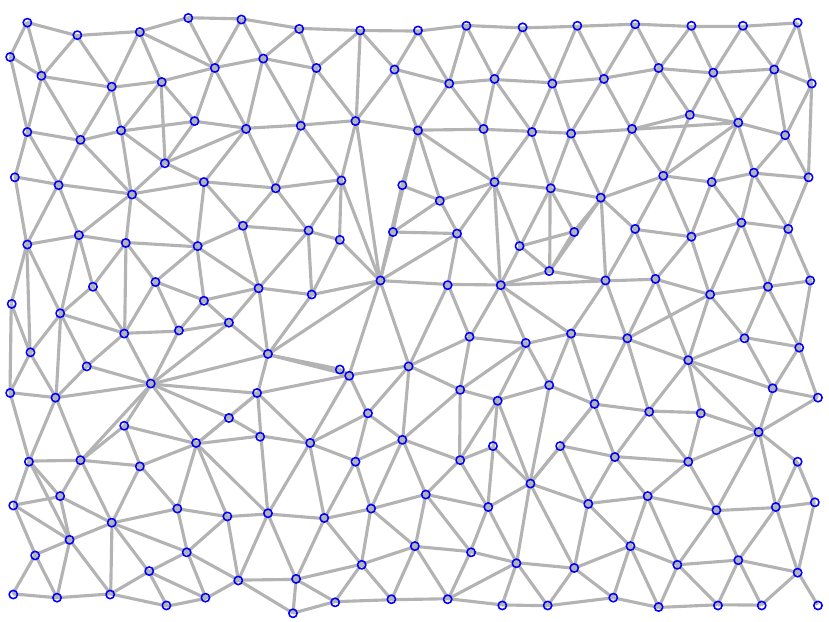}&
   \includegraphics[width=\wtab,height=\htab]{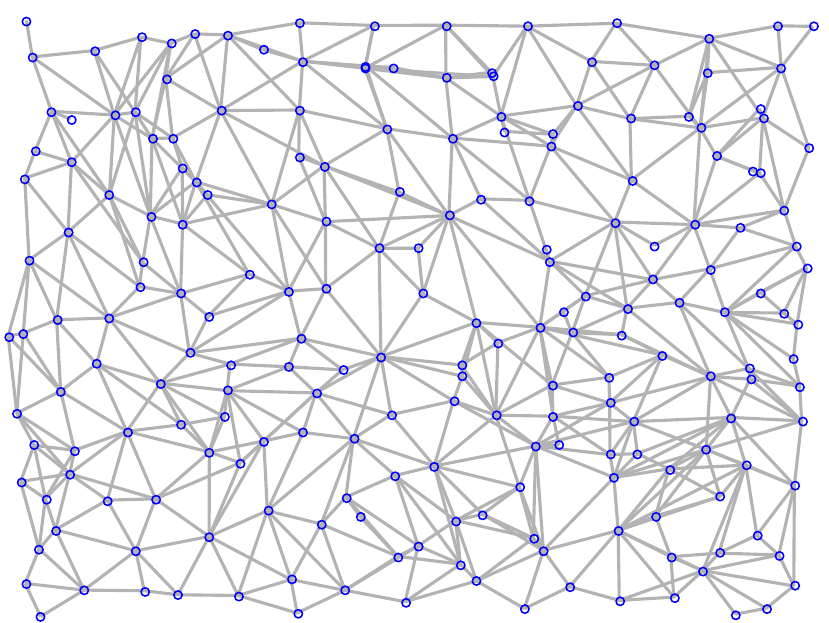}&
  \includegraphics[width=\wtab,height=\htab]{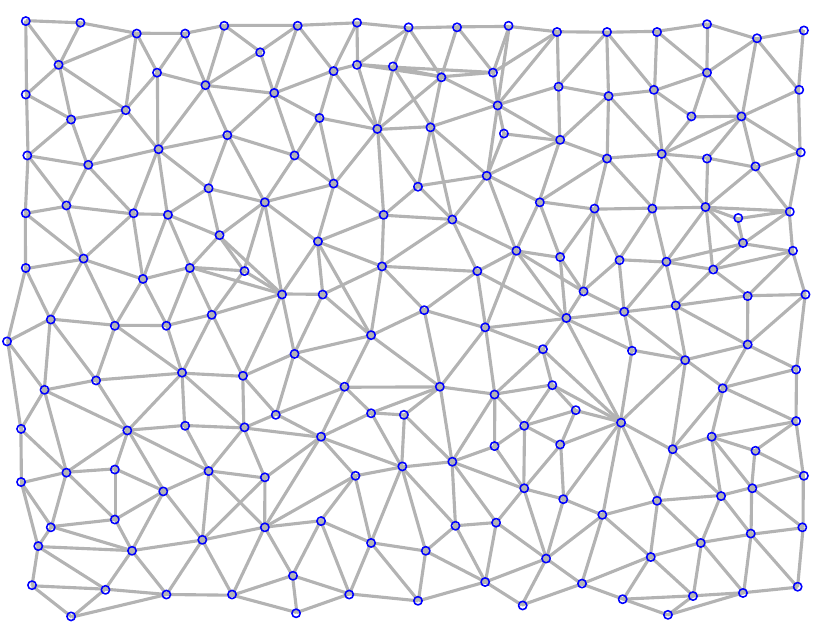}\\ 
   {\footnotesize SLIC \cite{achanta2012}}&
   {\footnotesize  ERGC \cite{buyssens2014}}&
   {\footnotesize  WP \cite{machairas2015}}&
   {\footnotesize  LSC \cite{li2015}}&
   {\footnotesize  SCALP \cite{giraud2016}}\\
  \end{tabular}
  }
  \end{center}
  \caption{
Decomposition of a BSD image (top) and associated Delaunay graph (bottom) with methods default settings for $K$=$200$ superpixels.
 } \vspace{-0.4cm}
  \label{fig:ex_methods}
  \end{figure*}

\subsection{Evaluation of Superpixel Methods}

 \newcommand{\stab}{6mm}
\newcommand{\stabb}{4mm}
\begin{table}[t]
  \begin{center}
  {\footnotesize
 \begin{tabular}{@{\hspace{2mm}}p{2cm}@{\hspace{\stab}}c@{\hspace{\stab}}c@{\hspace{\stab}}c@{\hspace{2mm}}}
   {\footnotesize Method} & {\footnotesize C }&  {\footnotesize SRC}  \\  \hline
{\footnotesize SLIC	\cite{achanta2012}}	&$0.438$ $\pm$ $0.111$ & $0.518$ $\pm$ $\mathbf{0.072}$	\\
 {\footnotesize ERGC	\cite{buyssens2014}}    &$0.367$ $\pm$ $0.040$ & $0.456$ $\pm$ $\mathbf{0.015}$ 	\\	
 {\footnotesize WP	\cite{machairas2015}}	&$0.483$ $\pm$ $0.076$ & $0.559$ $\pm$ $\mathbf{0.043}$	\\
 {\footnotesize LSC	\cite{li2015}	    }   &$0.228$ $\pm$ $0.046$ & $0.327$ $\pm$ $\mathbf{0.035}$ 	\\	
 {\footnotesize SCALP	\cite{giraud2016}}	&$0.515$ $\pm$ $0.115$ & $0.586$ $\pm$ $\mathbf{0.073}$ 	\\
  \end{tabular}
  }
    \end{center}
   \caption{
   Comparison of the superpixel methods regularity averaged on the BSD images for several superpixel scales $K=[50,1000]$.
  } \vspace{-0.4cm}
    \label{table:comp}
\end{table}

In this section, 
we consider state-of-the-art methods that enable to 
set a compactness parameter:
SLIC	\cite{achanta2012},	
ERGC	\cite{buyssens2014},	
Waterpixels (WP)	\cite{machairas2015},
LSC	\cite{li2015} 	and
SCALP	\cite{giraud2016}.
Decomposition examples with the associated Delaunay graph
are illustrated in Figure \ref{fig:ex_methods}.
In Table \ref{table:comp}, 
we demonstrate that SRC provides a more robust
regularity measure of superpixel methods.
We compute  decompositions at several
scales $K$ (from 50 to 1000 superpixels),  with the default compactness settings, and 
average the results on the 200 BSD images.
SRC is more robust to the superpixel scale, since it reports lower variance.

We also consider the standard
undersegmentation error (UE)
metric measuring the number of pixels that belong to several image objects, and
the boundary recall (BR), that measures the detection of ground truth contours (defined for instance in \cite{vandenbergh2012}).
The scale $K$ is set to 300 superpixels, 
and the UE and BR results are averaged on the BSD images,
decomposed with different compactness settings.
Hence, for each method, UE and BR results are computed on
several regularity levels.
We consider the optimal UE and BR of each method,
and the standard deviation between the corresponding C and SRC measures
are respectively $0.0871$ and $0.0803$ for UE, and
$0.0746$ and $0.0694$ for BR.
The reduced variance demonstrates the relevance of our metric, since
optimal decomposition performances of different methods,
evaluated with UE and BR, are obtained for more similar SRC.

        \newcommand{\htabm}{0.09\textwidth}
        \newcommand{\wtabm}{0.155\textwidth}
\begin{figure}[t!]
\begin{center}
{\footnotesize
 \begin{tabular}{@{\hspace{0mm}}c@{\hspace{1mm}}c@{\hspace{1mm}}c@{\hspace{0mm}}}
{ \scriptsize C = $0.296$ $|$ SRC = $0.434$}&
{ \scriptsize C = $0.401$ $|$ SRC = $0.562$}&
{ \scriptsize C = $0.371$ $|$ SRC = $0.570$}\\ \vspace{-0.1cm}
 \includegraphics[width=\wtabm,height=\htabm]{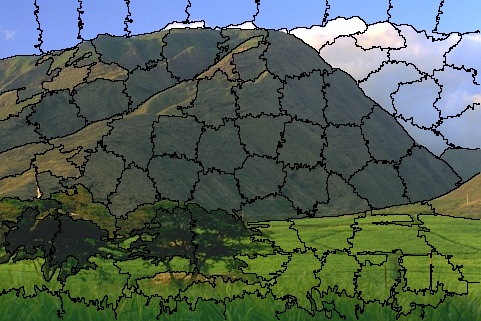}&
  \includegraphics[width=\wtabm,height=\htabm]{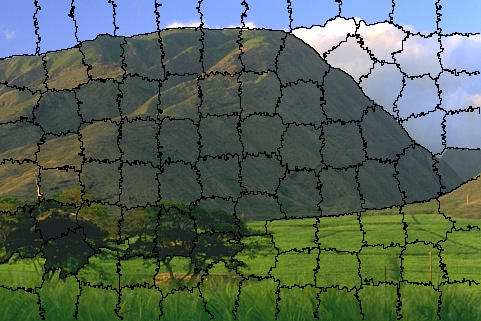}&
   \includegraphics[width=\wtabm,height=\htabm]{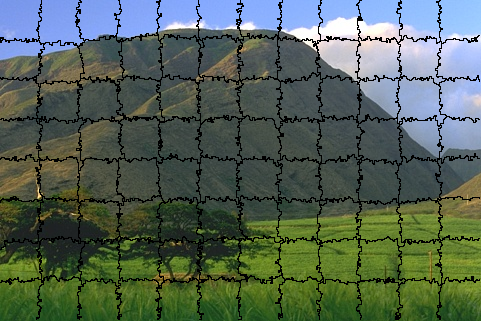}\\ 
   {\footnotesize $m=10$}&
   {\footnotesize $m=50$}&
   {\footnotesize $m=200$}
  \end{tabular}
  }
  \end{center}
  \caption{
 Example
 of  noisy superpixels computed from \cite{achanta2012} with $K$=100 superpixels, 
 for several compactness settings $m$.
  } \vspace{-0.2cm}
  \label{fig:slic_noisy}
  \end{figure}

\subsection{Robustness to Noise}

In this section, we further demonstrate the robustness of SRC to noisy boundaries.
We randomly perturb the trade-off between contour adherence and compactness of SLIC \cite{achanta2012}
to generate noisy decompositions (see Figure \ref{fig:slic_noisy}).
We compute decompositions for different values of compactness parameter $m$, 
and we report  the C and SRC measures averaged on the BSD 
in Figure \ref{fig:slic_noisy_regu}.
The circularity appears to be impacted by the noisy borders, 
to such an extent that it does not express the global shape regularity
that increases with $m$, since it drops from $m=75$.
Nevertheless, SRC is proportional to the 
evolution of the compactness parameter of \cite{achanta2012},
demonstrating that it better expresses the regularity of a decomposition,
and is not mainly sensitive to contour smoothness.

  \newcommand{\htabb}{0.19\textwidth}
    \newcommand{\wtabb}{0.4\textwidth}
  \begin{figure}[t!]
{\footnotesize
  \begin{center}
 \begin{tabular}{@{\hspace{0mm}}c@{\hspace{0mm}}}
 \includegraphics[width=\wtabb,height=\htabb]{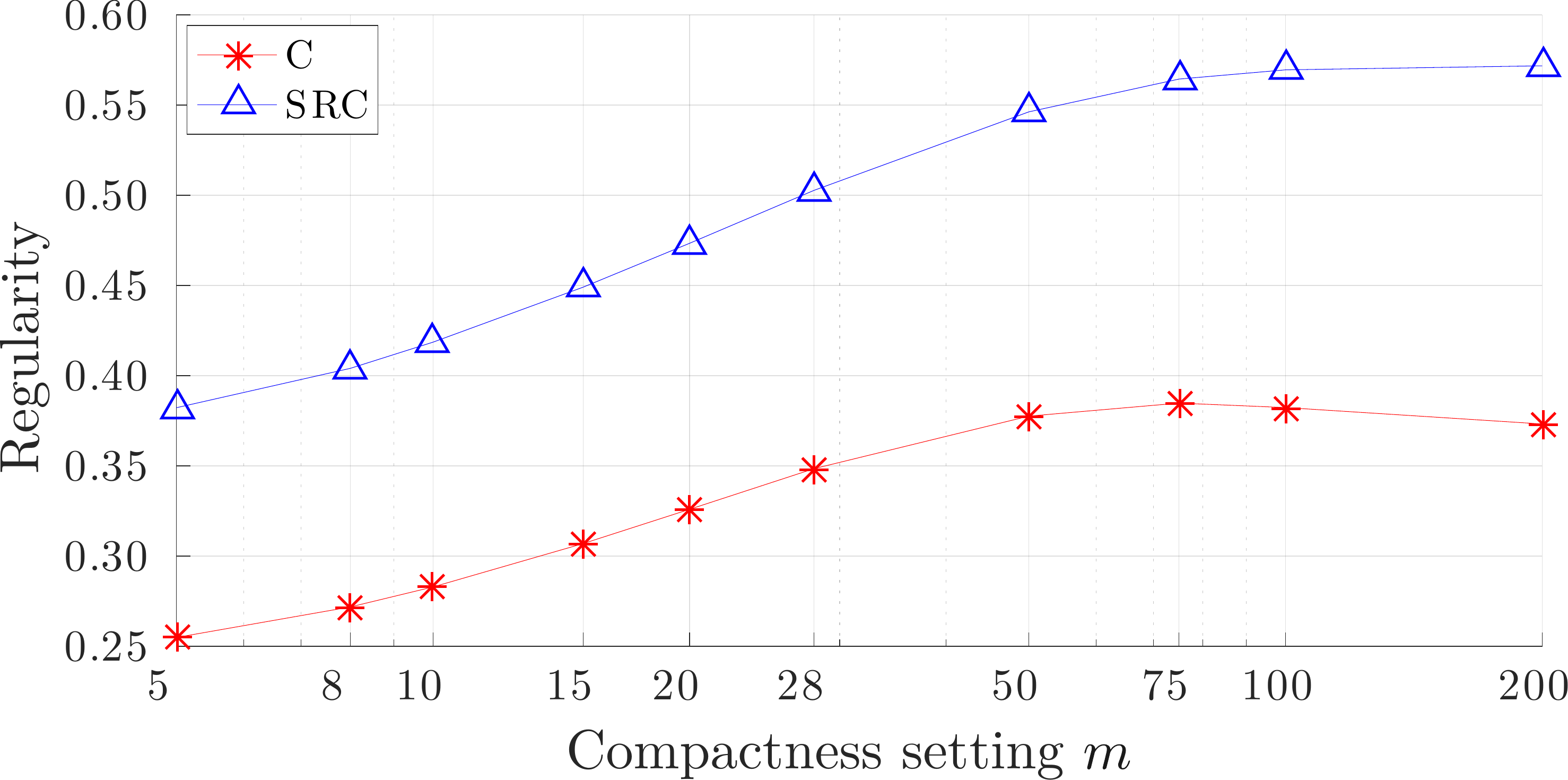}\\
  \end{tabular}
  \end{center}
}
  \caption{
  Regularity evolution of noisy superpixels computed from \cite{achanta2012} 
  for several compactness settings on the BSD.
  } \vspace{-0.35cm}
  \label{fig:slic_noisy_regu}
  \end{figure}

    \newcommand{\ttth}{0.12\textwidth}
  \newcommand{\tttt}{0.115\textwidth}
\begin{figure}[t!]
\begin{center}
{\footnotesize
\begin{tabular}{@{\hspace{0mm}}c@{\hspace{0.5mm}}c@{\hspace{0mm}}}
$\text{C}=0.741$ $|$ $\text{SRC}=0.790$ & $\text{C}=0.884$ $|$ $\text{SRC}=1.000$\\
  \includegraphics[width=\tttt,height=\ttth]{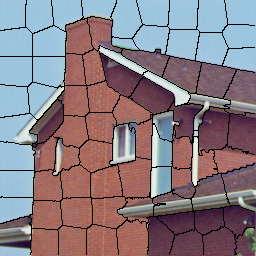}
    \includegraphics[width=\tttt,height=\ttth]{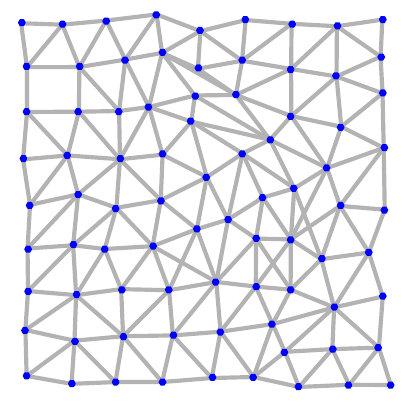} &
      \includegraphics[width=\tttt,height=\ttth]{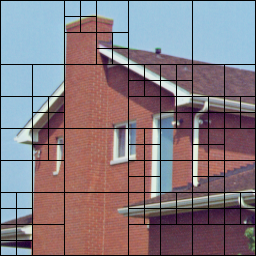}
    \includegraphics[width=\tttt,height=\ttth]{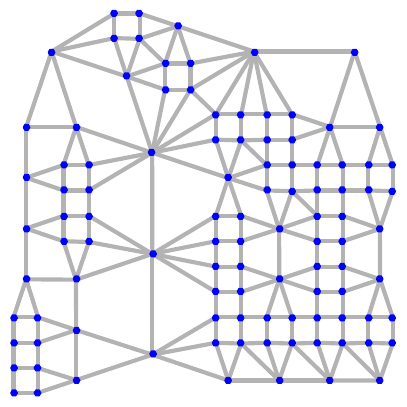} \\
     SLIC \cite{achanta2012} & Quadtree   \\
 \end{tabular}
  }
\end{center}
\caption{Example of SLIC \cite{achanta2012} and quadtree-based partitioning, with 
Delaunay graphs, connecting adjacent superpixel barycenters.} \vspace{-0.4cm}
\label{fig:quad}
\end{figure}

  \subsection{Global Regularity Evaluation}
  
  This work and \cite{schick2012}, that introduced circularity,
  focus on a local compactness definition, where  each superpixel is independently evaluated.
  Such local evaluation seems in line with
  the visual regularity of the graphs (see Figure \ref{fig:ex_methods}), but
  it does not consider
  the global size regularity within the whole decomposition. 
  Although most methods such as \cite{achanta2012,machairas2015,giraud2016} produce superpixels approximately containing the same number of pixels,
  other methods may produce  partitions of irregular sizes \cite{felzenszwalb2004,vandenbergh2012,rubio2016}.
  In Figure \ref{fig:quad}, 
  we represent an example of SLIC superpixels \cite{achanta2012} and standard
  quadtree partition, which produces larger squares in areas with lower color variance.
  Decompositions are shown with their associated Delaunay graph, connecting barycenters of adjacent superpixels.
  %
  Since it only produces square areas, the local regularity of such quadtree partition is high ($\text{SRC}=1$),
  although
  the associated graph shows inconsistent distances between the barycenters of connected superpixels,
  contrary to the one of \cite{achanta2012}.
  Numerical evaluation of graph regularity is a complex issue \cite{diestel1997}, and
  future works will investigate the global regularity notion and measure in the superpixel context.

  \section{Conclusion}
  
In this paper, we focus on the notion of regularity, \emph{i.e.}, 
compactness in the superpixel context.
We consider that a regular shape should verify these aspects:
convexity, balanced repartition and contour smoothness, and we
define a new metric that better expresses the local regularity, 
and is robust to scale and noise.
Most of decomposition methods tend to achieve a trade-off between
segmentation accuracy and shape regularity.
This work enables to relevantly compare 
superpixel algorithms
and
provides accurate regularity information on the decomposition inputs of
superpixel-based pipelines.
Nevertheless, local compactness measure does not express the global regularity notion,
which will be investigated in future works.

  \newpage
\label{sec:ref}
\bibliographystyle{IEEEbib}
\bibliography{ICIP_SP,refs}

\end{document}